\documentclass[letterpaper, 10 pt, journal, twoside]{IEEEtran}
\usepackage{amsmath,amsfonts}
\usepackage{algorithmic}
\usepackage{algorithm}
\usepackage{array}
\usepackage[caption=false,font=normalsize,labelfont=sf,textfont=sf]{subfig}
\usepackage{textcomp}
\usepackage{stfloats}
\usepackage{url}
\usepackage{verbatim}
\usepackage{graphicx}
\usepackage{cite}
\usepackage{cite}
\usepackage{indentfirst}
\usepackage{float}
\usepackage{epstopdf}
\usepackage{tikz}
\usepackage{comment}
\usepackage{amsmath,amssymb} 
\usepackage{color}
\usepackage[accsupp]{axessibility}  

\usepackage{booktabs}
\usepackage{multicol}
\usepackage{geometry} 
\newcommand{\reflabel}{dummy} 

\newcommand{\be}{\begin{equation}}
\newcommand{\ee}{\end{equation}}
\newcommand{\eqlabel}[1]{\label{eq:\reflabel-#1}}
\renewcommand{\eqref}[2][\reflabel]{(\ref{eq:#1-#2})}

\newcommand{\seclabel}[1]{\label{sec:\reflabel-#1}}
\newcommand{\secref}[2][\reflabel]{Section~\ref{sec:#1-#2}}

\newcommand{\figlabel}[2][\reflabel]{\label{fig:#1-#2}}
\newcommand{\figref}[2][\reflabel]{Fig.~\ref{fig:#1-#2}}

\newcommand{\tablelabel}[2][\reflabel]{\label{table:#1-#2}}
\newcommand{\tableref}[2][\reflabel]{Table~\ref{table:#1-#2}}

\newcommand{\fl}[1]{\textcolor{black}{#1}}
\newcommand{\rs}[1]{\textcolor{black}{#1}}

\newcommand{\point}{\textbf{x}}
\newcommand{\dir}{\textbf{d}}
\newcommand{\clrval}{c}
\newcommand{\clr}{C}

\newcommand{\dens}{\rho}
\newcommand{\mean}{\mu}
\newcommand{\uncert}{\sigma}

\newcommand{\rayclrval}{c_r}
\newcommand{\rayclr}{C_r}
\newcommand{\raysetclr}{C_I}
\newcommand{\raysetclrval}{c_I}

\newcommand{\numrayset}{R}
\newcommand{\param}{\mathbf{\theta}}
\newcommand{\nerf}{\mathbf{F_\theta}}

\usepackage{xspace}

\makeatletter
\DeclareRobustCommand\onedot{\futurelet\@let@token\@onedot}
\def\@onedot{\ifx\@let@token.\else.\null\fi\xspace}

\def\ie{\emph{i.e}\onedot} 
 
\def\etc{\emph{etc}\onedot} 
 
\def\etal{\emph{et al}\onedot}
\makeatother

\begin{document}
\begin{sloppypar}

\title{NeurAR: Neural Uncertainty for Autonomous 3D Reconstruction with Implicit Neural Representations}

\author{Yunlong Ran$^{1}$, Jing Zeng$^{1}$, Shibo He$^{2}$, Jiming Chen$^{2}$, Lincheng Li$^{3}$, Yingfeng Chen$^{3}$, Gimhee Lee$^{4}$, Qi Ye$^{2}$
\thanks{Manuscript received: August 7, 2022; Revised: November 23, 2022; Accepted: December 21, 2022. This paper was recommended for publication by Editor Cesar Cadena upon evaluation of the Associate Editor and Reviewers’ comments. This work was supported in part by NSFC under Grants 62233013, 62088101, 62103372, and the Fundamental Research Funds for the Central Universities, China.\textit{ (Corresponding Author: Qi Ye)} }
\thanks{$^{1}$Yunlong Ran and Jing Zeng are with the College of Control Science and Engineering, Zhejiang University, Hangzhou, China.
        {\tt\footnotesize \{yunlong\_ran, zengjing\}@zju.edu.cn}
        }%
\thanks{$^{2}$Shibo He, Jiming Chen, and Qi Ye are with the College of Control Science and Engineering, the State Key Laboratory of Industrial Control Technology, Zhejiang University, and the Key Laboratory of Collaborative Sensing and Autonomous Unmanned Systems of Zhejiang Province.
        {\tt\footnotesize  \{s18he, cjm  qi.ye\}@zju.edu.cn}
        }%
\thanks{$^{3}$Lincheng Li and Yingfeng Chen are with Fuxi AI Lab, NetEase Inc., Hangzhou, China.
        {\tt\footnotesize \{lilincheng, chenyingfeng1\}@corp.netease.com}
        }%
\thanks{$^{4}$Gimhee Lee is with the Department of Computer Science, National University of Singapore, Singapore.
        {\tt\footnotesize dcslgh@nus.edu.sg}
        }%

\thanks{Digital Object Identifier (DOI): see top of this page.}
}

\markboth{IEEE Robotics and Automation Letters. Preprint Version. Accepted January 2023}
{Ran \MakeLowercase{\textit{et al.}}: Neural Uncertainty for Autonomous 3D Reconstruction with Implicit Neural Representations} 


\maketitle

\begin{abstract}
Implicit neural representations have shown compelling results in offline 3D reconstruction and also recently demonstrated the potential for online SLAM systems. However, applying them to autonomous 3D reconstruction, where a robot is required to explore a scene and plan a view path for the reconstruction, has not been studied. In this paper, we explore for the first time the possibility of using implicit neural representations for autonomous 3D scene reconstruction by addressing two key challenges: 1) seeking a criterion to measure the quality of the candidate viewpoints for the view planning based on the new representations, and 2) learning the criterion from data that can generalize to different scenes instead of a hand-crafting one. \rs{To solve the challenges, firstly, a proxy of Peak Signal-to-Noise Ratio (PSNR) is proposed to quantify a viewpoint quality; secondly, the proxy is optimized jointly with the parameters of an implicit neural network for the scene. }
With the proposed view quality criterion from neural networks (termed as Neural Uncertainty), we can then apply implicit representations to autonomous 3D reconstruction. Our method demonstrates significant improvements on various metrics for the rendered image quality and the geometry quality of the reconstructed 3D models when compared with variants using TSDF or reconstruction without view planning. Project webpage https://kingteeloki-ran.github.io/NeurAR/ 

\end{abstract}



\section{Introduction}
\label{sec:intro}


\IEEEPARstart{A}{utonomous} 3D reconstruction has a wide range of applications, e.g. augmented/virtual reality, autonomous driving, filming, gaming, medicine, architecture. 
The problem requires a robot to make decisions about moving towards which viewpoint in each step to get the best reconstruction quality of an unknown scene with the lowest cost, i.e. view planning. In this work, we assume a robot can localize itself and at each viewpoint, the information of a scene is captured by an RGB image (with an optional depth image).

Implicit neural representations for 3D objects 
have shown their potential to be precise in geometry encoding, efficient in memory consumption (adaptive to scene size and complexity), predictive in filling unseen regions and flexible in the amount of training data.
The reconstruction with offline images~\cite{mildenhall2020nerf}
or online images with cameras held by human~\cite{sucar2021imap} has achieved compelling results recently with implicit neural representations. 
However, leveraging these advancements to achieve high-quality autonomous 3D reconstruction has not been studied.

Previous 3D representations for autonomous 3D reconstruction include point cloud, volume, and surface. To plan the view without global information of a scene, previous work resorts to a greedy strategy: given the current position of a robot and the reconstruction status, they quantify the quality of the candidate viewpoints via information gain to plan the next best view (NBV). In these works, the information gain relies on hand-crafted criteria, each designed ad-hoc for a particular combination of a 3D representation and a reconstruction algorithm. For example, Mendez \etal~\cite{mendez2017taking} define view cost by triangulated uncertainty given by the algorithm inferring depth from stereo RGB images, Isler \etal~ \cite{isler2016information} quantify the information gain for a view using entropy in voxels seen in this viewpoint, Wu \etal~\cite{wu2014quality} identify the quality of the view by a \rs{Poisson field} from point clouds and Song \etal~\cite{song2018surface} leverage mesh holes and boundaries to guide the view planning. 

\begin{figure*}
    \centering
    
    \includegraphics[ clip, page=1, width=0.8\linewidth]{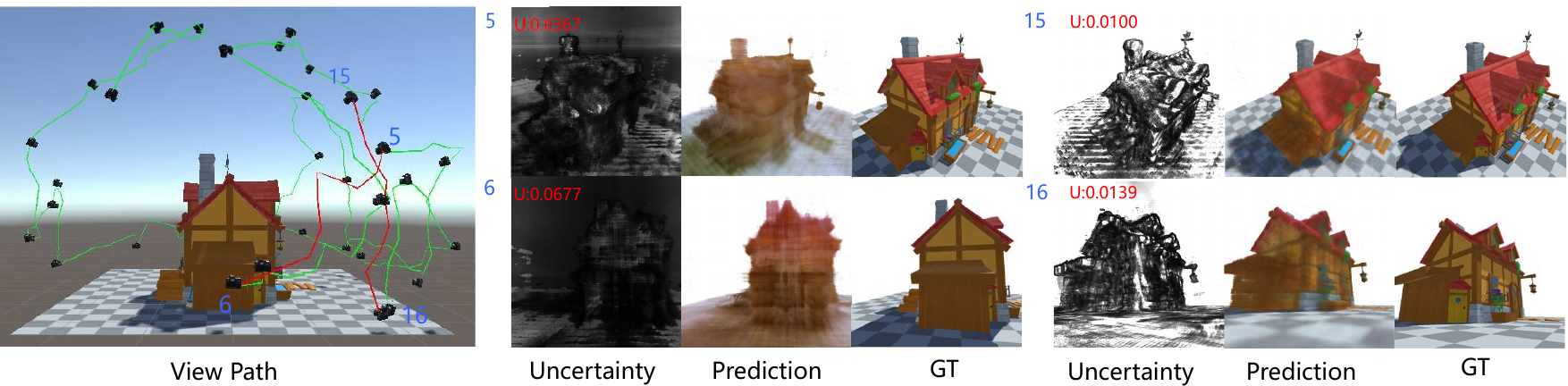}

    \caption{View paths planned by NeurAR and the uncertainty maps used to guide the planner. Given current pose 5 and 15,  paths are planned toward viewpoints having higher uncertainties, \ie pose 6 and  16. Uncertainties of the viewpoints are shown in red text. Notice that darker regions in uncertainty maps relate to worse quality regions of rendered images. 
    }
    \figlabel{planning}
    \vspace{-6mm}
\end{figure*}

To use implicit neural representations for autonomous 3D reconstruction, \rs{a key capability is to} quantify the quality of the candidate viewpoints. \textit{For implicit neural representations, how to define the viewpoint quality? Is it possible for the neural network to learn a measurement of the quality from data instead of defining it heuristically?} In this paper, we make efforts to answer both questions.

The quality of reconstructed 3D models can be measured by the quality of images rendered from different viewpoints and the measurement is adopted in many offline 3D reconstruction works. PSNR is one popular measurement and it is defined according to the difference between the images rendered from the reconstructed model and the ground truth model. 

The ground truth images, however, are unavailable for unvisited viewpoints to calculate PSNR during autonomous reconstruction. Is it possible to learn a proxy for PSNR? In \cite{PRML,ye2018occlusion}, \rs{the authors point out} that if the target variable to regress is under a Gaussian noise model and the target distribution conditioned on the input is optimized by maximum likelihood, the optimum value of the noise variance is given by the residual variance of the target values and the regressed ones. Inspired by this, we assume the color to regress for a spatial point in a scene as a random variable modeled by a Gaussian distribution.  The Gaussian distribution models the uncertainty of the reconstruction and the variance quantifies the uncertainty. When the regression network converges, the variance of the distribution is given by the squared error of the predicted color and the ground truth color; the integral of the uncertainty of points in the frustum of a viewpoint can be taken as a proxy of PSNR to measure the quality of candidate viewpoints.

With the key questions solved, we are able to build an autonomous 3D reconstruction system (NeurAR) using an implicit neural network. In summary, the contributions of the paper are:
\begin{itemize}

\item \rs{We propose the first autonomous 3D reconstruction system using an implicit neural representation. }
\item \rs{ We propose a novel view quality criterion that learns online from continuously added input images per target scene instead of hand-engineering or learning from a large corpus of 3D scenes.}
\item Our proposed method significantly improves on various metrics from alternatives using voxel-based representations or using man-designed paths for the reconstruction. 
\end{itemize}

\vspace{-1mm}
\section{Related Work}

\noindent\textbf{View Planning} Most view planning methods focus on the NBV problem which uses feedback from current partial reconstruction to determine NBV.  According to the representations for the 3D models, these methods can be divided into voxel-based method~\cite{isler2016information,mendez2017taking} and surface-based method~\cite{wu2014quality,schmid2020efficient}. 

The voxel-based methods are most commonly used due to their simplicity in representing space. Vasquez-Gomez \etal~\cite{vasquez2017view} analyze a set of boundary voxels and determine NBV for dense 3D modeling of small-scale objects. Stefan Isler \etal~\cite{isler2016information} provide several metrics to quantify the volume information contained in the voxels. Mendez \etal~\cite{mendez2017taking} define the information gain by the triangulation uncertainty of stereo images. Despite the simplicity, these methods suffer from memory consumption with growing scene complexity and higher spatial resolutions. 

A complete volumetric map does not necessarily guarantee a perfect 3D surface. Therefore, researchers propose to analyze the shape and quality of the reconstructed surface for NBV~\cite{wu2014quality,schmid2020efficient}.
Wu  \etal~\cite{wu2014quality} estimate a confidence map representing the completeness and smoothness of the constructed Poisson isosurface and the confidence map is used to guide the calculation of NBV. Schmid  \etal~\cite{schmid2020efficient} propose information gain to evaluate the quality of observed surfaces and unknown voxels near the observed surfaces, then plan a path by RRT. 

\rs{In contrast, some methods explore function learning to solve the NBV problem ~\cite{devrim2017reinforcement,schmid2022fast,hepp2018learn}.  Supervised learning-based methods~\cite{hepp2018learn} learn \fl{information gain of a viewpoint given a partial occupancy map} and ~\cite{schmid2022fast} learns \fl{an informed distribution of high-utility viewpoints based on a partial occupancy map}. For the reinforcement learning-based methods~\cite{devrim2017reinforcement}, no hand-crafted heuristics are required and an agent explores the viewpoints with high overall coverage.  These learning-based methods require a large-scale dataset for training and may be hard to generalize to different scenes. }





\rs{Different from the hand-crafted
heuristics, the learned NBV policy, and the existing learning-based information gain, our proposed neural uncertainty is learned per target scene during the reconstruction, requiring no manual definition, no large training set, and being able to work in any new scene.}

\noindent\textbf{Online Dense Reconstruction} \rs{For online 3D reconstruction from RGB images, most methods use Multi-View-Stereo (MVS)~\cite{schonberger2016pixelwise} to reconstruct dense models by first getting a sparse set of initial matches, iteratively expanding matches to nearby locations and performing surface reconstruction. With the release of commodity RGB-D sensors, the fusion of point clouds reprojected from depth images gains popularity. KinectFusion ~\cite{izadi2011kinectfusion} achieves real-time 3D reconstruction with a moving depth camera by integrating points from depth images with Truncated Signed Distance Functions (TSDFs). OctoMap ~\cite{hornung2013octomap} builds a probabilistic occupancy volume based on octree. Recently, implicit representations have shown compelling results in 3D reconstruction, either for the radiance field approximation from RGB images~\cite{nerf} or the shape approximation from point clouds ~\cite{gropp2020implicit}. The novel representations are also studied for online dense reconstruction. iMAP~\cite{sucar2021imap} adopts MLPs as the scene representation and reconstructs the scene from RGBD images. In addition to using MLPs to represent a scene implicitly,  NeRFusion ~\cite{zhang2022nerfusion
} further combines a feature volume to fuse information from different views as a latent scene representation.
Similarly, we use the implicit neural function to represent 3D models but we focus on how to leverage this representation for autonomous view planning.
}




\begin{figure}[htbp]
\vspace{-2mm}
\centering
\includegraphics[ width=0.9\linewidth]{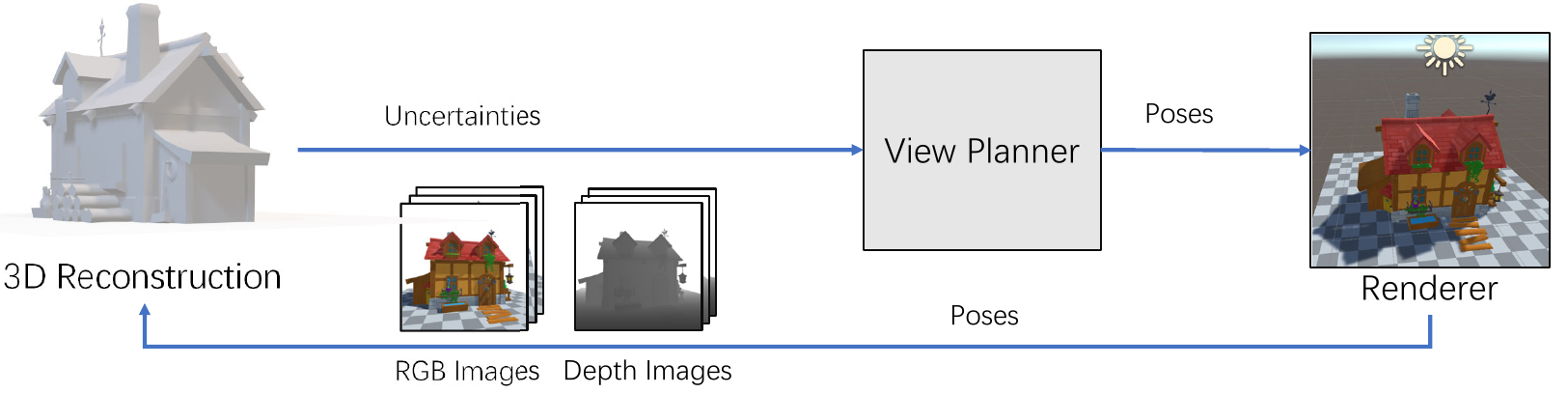}
\caption{The pipeline of our method}
\figlabel{pipeline}
\vspace{-5mm}
\end{figure}

\vspace{-1mm}
\section{Method}
\label{sec:method}
\subsection{Problem Description and System Overview}
\seclabel{overview}

The problem considered is to generate a trajectory for a robot that yields high-quality 3D models of a bounded priori unknown scene and fulfills robot constraints.  The trajectory is defined as a sequence of viewpoints $V=(v_1,...,v_n)$ where $v_i \in \mathcal{R}^3 \times \mathcal{SO}(2)$ (usually roll is not considered). Let $S$ be the set of all the sequences. The problem can therefore be expressed as 
\begin{equation}
\centering
\begin{aligned}
V^* = \mathop{\arg\max}\limits_{V \in S} I(V) \text{ s.t. } L(V) \leq L_{max},
\end{aligned}
\eqlabel{opttraj}
\end{equation}
where $I(V)$ equals $\Sigma I(v_i)$, an objective function measuring the contributions from different viewpoints to the reconstruction quality and $ L(V)$ is the length of the trajectory.

One difficulty of the problem is defining the objective function itself as there is no ground truth data to measure the quality of the reconstruction. Hence, one major line of work in this field is seeking a surrogate quantity that can be used for the view sequence optimization, which is also the focus of this paper.

Finding the best sequence for \eqref{opttraj} is prohibitively expensive and most work resolves this problem by a greedy strategy. At each step of the reconstruction process, a number of candidate viewpoints are sampled according to some constraints, the contributions of these viewpoints to the reconstruction quality are evaluated based on the current reconstructed scene and the viewpoint with the most contribution is chosen as NBV. The process is repeated until a maximum length is reached. To move a robot to NBV, various path planning methods can be applied to navigate the robot. 

\rs{In this work, we follow the greedy strategy and choose an existing RRT* based view planner. For each viewpoint, an RGB image (with an optional depth image) is captured. To achieve an autonomous 3D reconstruction system depending only on the input images, joint estimation of camera poses and the target scene is required. To focus on view planning with implicit scene representations, we assume the ground truth camera poses are given and leave the localization of the camera as future work.}

Accordingly, our proposed autonomous 3D reconstruction system can be divided into three modules (shown in the top of \figref{planning}): a 3D reconstruction module, a view planner module, and a \rs{simulation} module. At each step, the 3D reconstruction module reconstructs a scene represented by an implicit neural network with new images from the \rs{simulation} module and provides a quantity (neural uncertainty) measuring the reconstructed quality of each position (\secref{uncert}). To adapt the implicit neural representation to online reconstruction, we further propose several strategies for online training and acceleration in \secref{online}. The view planner module (\secref{planner}) samples viewpoints from empty space, measures the contributions of these sampled viewpoints by composing the neural uncertainty, chooses NBV and plans a view path to NBV. The \rs{simulation} module, built upon Unity Engine, renders RGBD images from new viewpoints and provides both the images and the camera poses to the 3D reconstruction module. 

\subsection{3D Reconstruction with Neural Uncertainty}
\seclabel{uncert}

In the section, we first formulate Neural Uncertainty when a target scene is approximated by an implicit neural network. Then we show how to optimize the uncertainty together with the network parameters. With the neural uncertainty formulation, a strong linear relationship is established between the uncertainty and the image quality metric, i.e. PSNR and therefore a proxy for PSNR is proposed. 

\subsubsection{Problem formulation}
\seclabel{formulation}

NeRF~\cite{mildenhall2020nerf} represents a continuous scene with an implicit neural function by taking the location of a point $\point$ on a ray of direction $\dir$ as inputs and its color value $\clrval$, density $\dens$ as outputs, \ie $\nerf(\point, \dir)=( \clrval, \dens)$. The function $\nerf$ is represented by an MLP. 

To learn the neural uncertainty for view planning, we treat the color as a random variable under a Gaussian distribution instead of taking it as a deterministic one. The color distribution can be represented as $p(\clr) = \mathcal{N}(\mean, \uncert^2)$, where the RGB channels of $\clr$ share the same $\uncert$. To estimate the distribution, we map the inputs  $(\point, \dir)$ to its parameters $(\mean, \uncert)$ of the distribution for the color variable $\clr$; in other words, the outputs of the MLP is $\mean$, $\uncert$ rather than $\clrval$, so we have  $\nerf(\point, \dir)=(\mean, \uncert, \dens)$.
Having defined the color distribution for a point $\point$ on a ray $\dir$,  we now deduce the rendered color distribution for a camera ray. Following NeRF, we define the rendered color as
\begin{equation}
\rayclr = \sum_{i=1}^{N} \omega_i \clr_i,
\eqlabel{calccolor}
\end{equation}
where $\omega_i$ is $o_i \prod_{j=1}^{i-1}(1-o_j)$, $o_i = (1-\exp(-\dens_i \delta_i))$ and $\delta_i = d_{i+1} - d_i$ which represents the inter-sample distance. $p(\clr_i)=\mathcal{N}(\mean_i, \uncert_i^2)$ is the color distribution for a point on the ray. Assuming \{$\clr_i$\} for points on the ray are independent from each other, the distribution for $\rayclr$ is a Gaussian distribution. The probability of the rendered color value being $\rayclrval$ is $p(\rayclr=\rayclrval)=\mathcal{N}(\rayclrval|\mean_r, \uncert_r)$, where the mean and variance are $\mean_r = \sum_{i=1}^{N} \omega_i \mean_i $ and $ \uncert_r^2 = \sum_{i=1}^{N} \omega_i \uncert_i^2$.

Similarly, assuming \{$\rayclr$\} for different rays are independent, the random variable $\raysetclr$ for the mean rendered color of rays sampled from an image pool is also under a Gaussian distribution, whose mean and variance are
\be
\mean_I = \frac 1 {\numrayset}\sum_{r=1}^{\numrayset}\mean_r = \frac 1 {\numrayset } \sum_{r=1}^{\numrayset} \sum_{i=1}^{N} \omega_{ri} \mean_{ri},
\eqlabel{meanimg}
\end{equation}
\be
\uncert_I^2=\frac 1 {\numrayset}  \sum_{r=1}^{\numrayset} \uncert_r^2=\frac 1 {\numrayset } \sum_{r=1}^{\numrayset}\sum_{i=1}^{N}\omega_{ri} \uncert_{ri}^2,
\eqlabel{uncertimg}
\end{equation}
where $r $ represents a camera ray tracing through a pixel from an image pool and $R$ the number of rays sampled for the image pool.

\subsubsection{Optimization}
\seclabel{Optimization}
We now consider how to optimize the network parameters $\param$ and determine $\mean, \sigma, \dens$ for each point on a ray (we ignore the subscript $ri$ for brevity). Consider a pool of images, the likelihood for the mean color value $\raysetclrval$ of a set of rays shooting from the pixels sampled from the image pool are obtained as $p(\raysetclr=\raysetclrval) = \mathcal{N}(\raysetclrval|\mean_I, \uncert_I)$. 
Taking the negative logarithm of the likelihood and ignoring the constant, we have 
\be
L_{color}=\log{\uncert_I} + \frac{L_{mean}}{2\uncert_I^2} \leq \log{\uncert_I} + \frac{L_I}{2\uncert_I^2},
\eqlabel{logloss}
\end{equation}
\be
L_{mean} = \lVert \raysetclrval - \mean_I \rVert_2^2 \leq  L_{I}=\frac 1 {\numrayset}  \sum_{r=1}^{\numrayset} \lVert \rayclrval - \mean_r \rVert_2^2.
\eqlabel{mseloss}
\end{equation}
For $L_{mean}$ of \eqref{mseloss}, the constraint for each pixel is too weak and the network is not able to converge a meaningful result (PSNR about 10). As we have supervision for the color of each pixel and $L_{mean}$ is smaller than or equal to $L_I$, we choose to minimize $L_I$ instead. The loss function above is differentiable w.r.t $\param$ and $\mean, \sigma, \dens$ are the outputs of MLPs ($\nerf$); we can use gradient descent to determine the network parameters $\param$ and $\mean, \sigma, \dens$. 
To make $\sigma$ positive, we let the network estimate  $\sigma^2$ and the output of MLPs $s$ is activated by $e^s$ to get $\sigma^2$.


Consider the minimization with respect to $\param$. Given $\sigma_I$, we can see the minimization of the maximum likelihood under a conditional Gaussian distribution for each point is equivalent to minimizing a mean-of-squares error function given by $L_I$ in \eqref{mseloss}. Apply $\nerf$ on a point on a ray and $\mean, \dens, \sigma$ can be obtained.

\subsubsection{Neural Uncertainty and PSNR}

For the variance $\sigma_I^2$, or Neural Uncertainty, the optimum value can be achieved by setting the derivative of $L_{color}$ with respect to $\sigma_I$ to zero, giving
\be
\sigma_I^2 =  L_{I}.
\eqlabel{relation}
\end{equation}
The equation above indicates that the optimal solution for $\sigma_I^2$ is the squared errors between the predicted image and the ground truth image. 

On the other hand, PSNR is defined as $10\log_{10}{\frac{MAX_I^2}{MSE}}$, where $MAX_I$ is the maximum possible pixel value of the image. When a pixel is represented using 8 bits, it is 255 and MSE is the mean squared error between two images, the same as $L_I$. Then we establish a linear relationship between the logarithm of Neural Uncertainty  and PSNR, i.e.
\be
PSNR =  A\log{\sigma}_I^2,
\end{equation}
where $A$ is a constant coefficient.

To verify the linear relationship, we scatter data pairs of $(PSNR,\log{\sigma_I^2})$ for images in the testing set evaluated at different iterations when optimizing $\nerf$ for a cabin scene (the scene is shown in \secref{results}). \rs{Two different training strategies are conducted: online training with images captured along a planned trajectory added sequentially and offline batch training using all the images precaptured from the trajectory}. As can be seen \figref{optimization}(c-d), a strong correlation exists between $PSNR$ and $\log{\sigma^2}$, whose Pearson Correlation Coefficient (PCC) is -0.96 for online and -0.92 for offline. The two variables are almost perfectly negatively linearly related. During the training, $\sigma_I$ and $L_I$ are jointly optimized. The loss curves of the uncertainty part $\log{\uncert_I}$ and the ratio part $ \frac{L_{I}}{\uncert_I^2}$ in \eqref{logloss} for online are shown in \figref{optimization}(b). Notice that the loss curve for the ratio part stays almost constant during training, which also verifies the effectiveness using uncertainty as a proxy of PSNR. 

\rs{For the verification and the usage of the linear relationship for autonomous reconstruction, we adopt NeRF~\cite{mildenhall2020nerf} as our implicit representation for a scene. From the formulation and the derivation of the relation between neural uncertainty and PSNR, our neural uncertainty is agnostic to the underlying function $\nerf$, which can be MLPs like NeRF or networks based on trainable feature vectors with MLP Decoder ~\cite{zhang2022nerfusion}}

\begin{figure}
     \centering     \includegraphics[ width=0.85\linewidth]{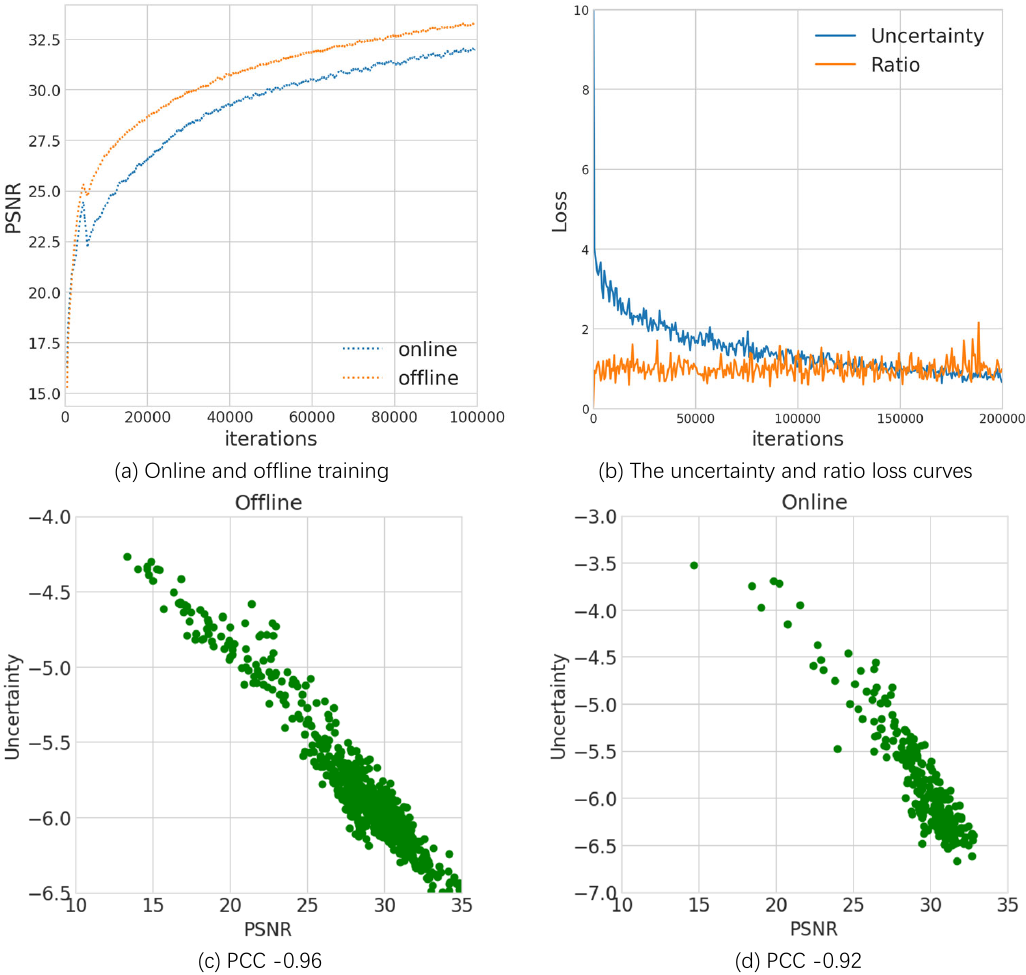}
     \caption{\rs{Loss curves and linear relationship between $\log{\sigma_I^2}$ and PSNR. PCC value of -1 signifies strong negative correlation. (a) Training loss curves for online training and offline training. (b) The loss curves for the uncertainty part $\log{\uncert_I}$ and the ratio part $ \frac{L_{I}}{\uncert_I^2}$ in \eqref{logloss}. (c) Linearity when the scene is optimized using offline images. (d) Linearity when the scene is optimized using online images.} }
\figlabel{optimization}
\vspace{-3mm}
\end{figure}

\subsection{Online Training and Acceleration}
\seclabel{online}
Though online reconstruction with Neural Uncertainty supervised by images can achieve similar accuracy with NeRF, the convergence is too slow for view planning. We accelerate training by introducing particle filter, depth supervision and a keyframe strategy. 

 \noindent \rs{\textbf{Particle filter}  keeps particles (rays) active in high loss region, which helps the network optimize details faster. At each step, when a new image is added to an image pool for the training, a set of particles are randomly sampled from the image. After an iteration of training, a quarter of particles are resampled according to the weight of particles, which is defined according to the loss of a ray, and the other particles are uniformly sampled from the image.} 

\rs{Particle filter is applied after coarse learning of the whole scene is done. At the early stage of training, the model knows little information about the scene and tends to have a higher loss for rays shooting at objects than that for empty space. This results in particles always staying at the surfaces of objects and therefore the network learns the whole space slower. Considering this, at the beginning iterations of the training, we use random sampling.}

\noindent \textbf{Depth supervision} can greatly speed up training. Depth images for NeRF are rendered similar to color images in \cite{mildenhall2020nerf}. We define depth loss and our final loss as 
\be
L_{depth} =  \frac 1 {\numrayset}\sum_{r=1}^{\numrayset} \lVert\hat{z}_r- z_r \rVert_2^2
\text{ , }\text{ }
L = L_{color} + \lambda_d L_{depth},
\eqlabel{finalloss}
\end{equation}
where $\hat{z}_r$ and $z_r$ represent the rendered depth from reconstructed 3D model $\nerf$ and the ground truth depth for a pixel. As depth captured from real sensors typically has noise, we find using depth supervision may make the model not able to converge well due to the conflict between noisy depth and the depth inferred from multiview RGB images. A balance needs to be made between the two cases. 
We strengthen depth supervision at the early stage of training to accelerate training and decrease it after getting a coarse 3D structure. The weight of depth loss is decreased from 1 to 1/10 after $N_d$ iterations to emphasize structure from multiview images. 

\noindent \textbf{Keyframe pool} We follow iMAP \cite{sucar2021imap} to maintain a keyframe pool containing 4 images for continual training. The pool is initialized with the first four views and during training,  the image with minimum image loss in the pool is replaced with a new image or an image from a seen images set. 

\subsection{View Planning with Neural Uncertainty}
\seclabel{planner}
As explained in \secref{overview}, we follow the greedy strategy. For the view path planning, we exploit the scenic path planner based on RRTs* \cite{mendez2017taking} to evaluate the efficacy of the proposed Neural Uncertainty in guiding the view planning. \rs{In the scenic path planner, the RRTs* algorithm samples from a prior distribution for the view quality cost-space approximated by Sequential Monte-Carlo (SMC) instead from $SE(3)$, which baises the growth of the tree towards areas with good NBV cost while also aims at a shortest Euclidean path. } 
Our view planner replaces their view cost with 
\be
C_{view} =\sigma_I^2,
\eqlabel{costv}
\end{equation}
where $\sigma_I^2$ is defined in \eqref{uncertimg} and here the image pool contains only one image.

For the view planning, we introduce a prior of objects in the center of the space for the candidate viewpoint sampling: the camera moves in a band with the nearest distance to the object $D_{near}$ meters and the farthest distance $D_{far}$ meters; the camera looks at directions pointing roughly to the center of the object; candidate viewpoints are sampled in a sphere with the center at the current camera position and the radius $R_{sample}$ meters.


\section{Results}
\seclabel{results}

\noindent\textbf{Data} The experiments are conducted on five 3D models, $drums$ from NeRF synthetic dataset, $Alexander$\footnote{https://3dwarehouse.sketchup.com/},  $cabin$, $tank$ and $monsters$  we collect online. The scenes used in the paper are mostly of the size $5m\times 5m \times 5m$,  and the larger scene $Alexander$ is about $80m\times 80m \times 80m$. $D_{near}$, $D_{far}$, $R_{sample}$ are 80m, 90m, 80m for $Alexander$ and 3m, 4m, 3m for other scenes.
RGBD images are rendered by Unity Engine and we assume their corresponding camera poses are known. To simulate depth noise,  all rendered depth images are added with noise scaling approximately quadratically with depth $z$ \cite{nguyen2012modeling}. The depth noise model is $ \epsilon = N(\mu(z),\sigma(z)) $ where $\mu(z) = 0.0001125z^2 + 0.0048875$ , $\sigma(z) = 0.002925z^2 + 0.003325 $ and the constant parameters are acquired by fitting the model to the noise reported for Intel Realsense L515 \footnote{https://www.intelrealsense.com/lidar-camera-l515/}. The depth noise model for $Alexander$ of large size, $\mu(z) = 0.00001235z^2 + 0.00004651$ , $\sigma(z) = 0.00001228z^2 + 0.00001571 $, the constant parameters acquired by fitting the model to the noise reported for Lidar VLP16\footnote{https://usermanual.wiki/Pdf/VLP16Manual.1719942037/view/}.

\noindent\textbf{Implementation details} The networks and the hyperparameters of all the experiments for different scenes below are set to the same value. Most hyper parameters use the default setting in NeRF, including parameters of Adam optimizer (with hyperparameters $\beta_1 = 0.9$,
$\beta_2 = 0.999$, $\epsilon = 10^{-7}$), 64/128 sampling points on a ray for coarse/fine sampling, a batch size of 1024 \etc. 
\begin{figure}[htbp]
\centering
\includegraphics[width=0.4\textwidth]{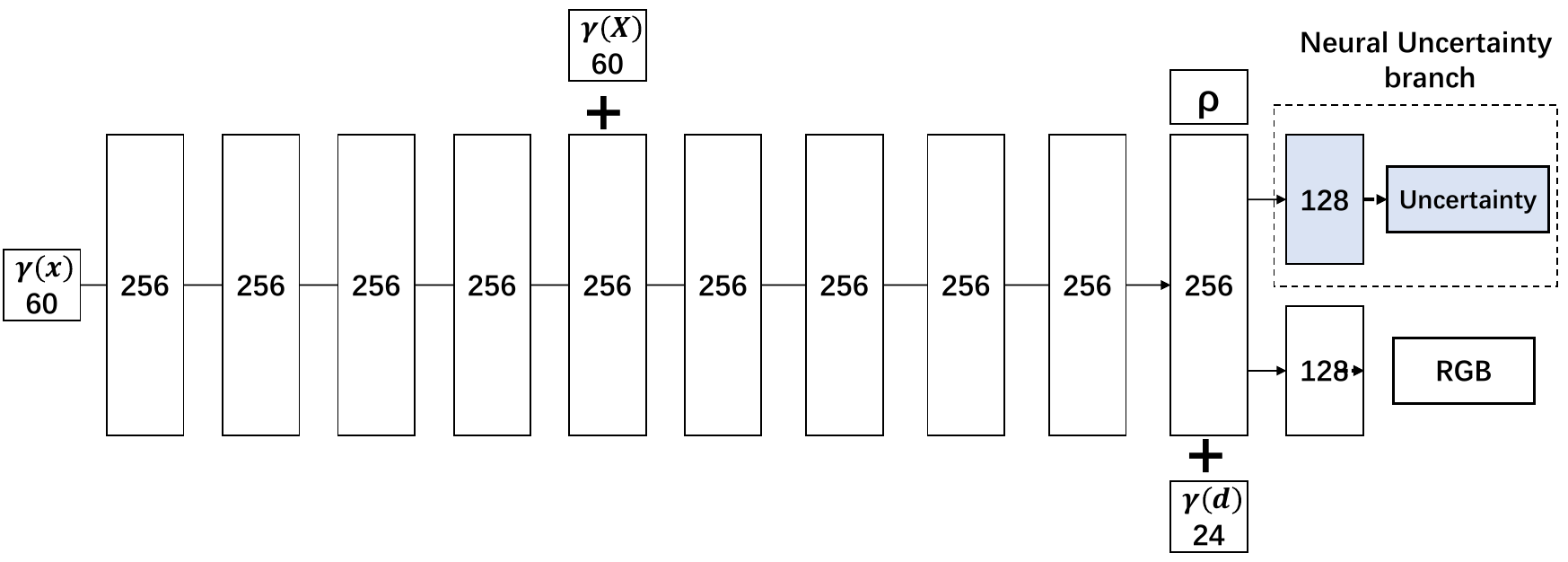}
\caption{The network architecture of the 3D reconstruction module of NeurAR. Similar to NeRF. The only difference is an additional branch of a 128 fully connected layer for the estimation of uncertainty}
\figlabel{arc}
\end{figure}

\begin{figure}[htbp]
  \centering
  \includegraphics[width=0.85\linewidth]{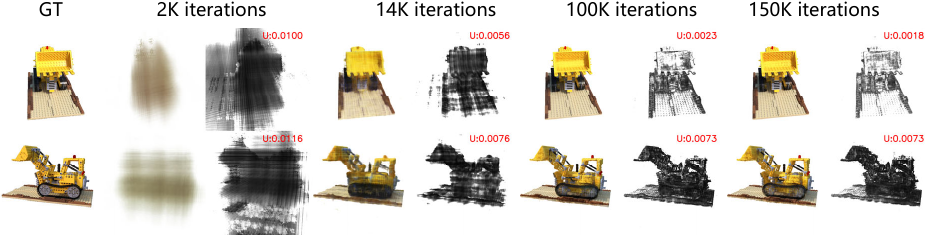}  
  \caption{Uncertainty maps from different viewpoints and training iterations. Top: Uncertainty maps for seen viewpoints; Bottom: uncertainty maps for an unseen viewpoint. The overall uncertainty decreases with training while the uncertainty for the unseen viewpoint stays high even when the network converges. }
  \figlabel{lego}
\end{figure}

\begin{figure*}[!ht]
  \setlength{\abovecaptionskip}{0.32cm}
  \centering
  \hspace{0cm}\includegraphics[width=0.85\linewidth]{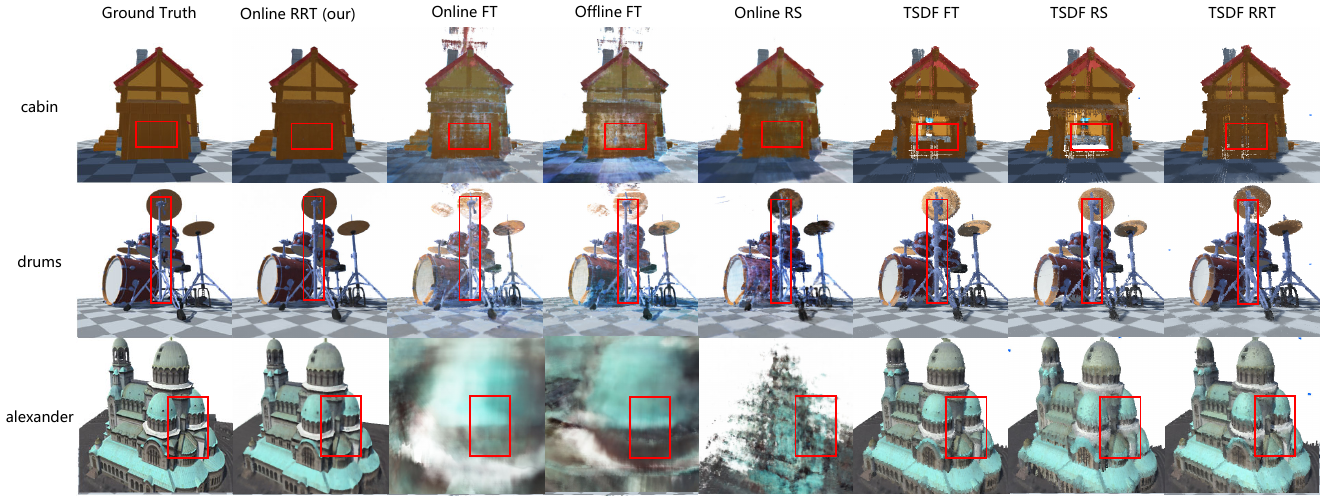}
  \caption{Comparison of the reconstruction models with different methods. Refer to the supplementary video for higher resolution, more comparison and more viewpoints. }
  \figlabel{compare_view}
  \vspace{-2mm}
\end{figure*}
\begin{table*}[!ht]
    \vspace{-2mm}
    \caption{Evaluations on the reconstructed 3D models using different methods }
    \tablelabel{tablepsnr}
    \centering
    
    \resizebox{\textwidth}{30mm}{
    \normalsize
    \setlength{\tabcolsep}{2.5mm}{
    
    \begin{tabular}{c|ccc|ccc|ccc|ccc|ccc}
    \toprule
         & \multicolumn{3}{c}{cabin} & \multicolumn{3}{c}{drums} & \multicolumn{3}{c}{alexander} & \multicolumn{3}{c}{tank} & \multicolumn{3}{c}{monsters}  \\ 
        Method & PSNR↑ & SSIM↑ & LPIPS↓ & PSNR↑ & SSIM↑ & LPIPS↓ & PSNR↑ & SSIM↑ & LPIPS↓ & PSNR↑ & SSIM↑ & LPIPS↓ & PSNR↑ & SSIM↑ & LPIPS↓  \\ \midrule
        TSDF FT & 21.17 & 0.768 & 0.140 & 18.67 & 0.746 & 0.150 & 19.30 & 0.662 & 0.190 & 20.39 & 0.782 & 0.151 & 20.42 & 0.786 & 0.126  \\ 
        TSDF RS & 20.87 & 0.75 & 0.151 & 18.68 & 0.738 & 0.160 & 19.15 & 0.649 & 0.195 & 18.64 & 0.749 & 0.171 & 20.19 & 0.777 & 0.131  \\
        TSDF RRT\cite{schmid2020efficient} & 20.34 & 0.739 & 0.171 & 17.95 & 0.716 & 0.181 & 18.86 & 0.644 & 0.199 & 21.36 & 0.772 & 0.172 & 20.14 & 0.773 &0.146  \\ 
        Offline FT & 24.31 & 0.842 & 0.108 & 21.17 & 0.831 & 0.117 & 8.40 & 0.431 & 0.606 & 21.71 & 0.819 & 0.142 & 25.01 & 0.880 & 0.076  \\ 
        Online FT & 23.42 & 0.833 & 0.114 & 21.49 & 0.831 & 0.117 & 8.71 & 0.419 & 0.597 & 23.03 & 0.839 & 0.124 & 24.42 & 0.878 & 0.077  \\ 
        Online RS & 26.74 & 0.864 & 0.082 & 23.66 & 0.859 & 0.0838 & 13.08 & 0.533 & 0.373 & 21.97 & 0.789 & 0.162 & 24.68 & 0.871 & 0.063  \\
        \rs{Online even cover.} & \rs{26.56} & \rs{0.868} & \rs{0.106} & \rs{24.82} & \rs{0.913} & \rs{0.049} & \rs{\textbf{24.30}} & \rs{\textbf{0.767}} & \rs{\textbf{0.182}} & \rs{24.81} & \rs{0.875} & \rs{0.108} & \rs{25.90} & \rs{0.909} & \rs{0.065}  \\
        \rs{Ours Offline} & \rs{\textbf{28.86}} & \rs{\textbf{0.917}} & \rs{\textbf{0.048}} & \rs{\textbf{26.45}} & \rs{\textbf{0.916}} & \rs{\textbf{0.051}} & \rs{23.98} & \rs{0.758} & \rs{0.188} & \rs{\textbf{27.54}} & \rs{\textbf{0.909}} & \rs{\textbf{0.064}} & \rs{\textbf{27.57}} & \rs{\textbf{0.927}} & \rs{\textbf{0.039}}  \\        
        Ours & 28.35 & 0.902 & 0.062 & 25.73 & 0.905 & 0.058 & 24.07 & 0.757 & 0.199 & 25.83 & 0.874 & 0.097 & 26.57 & 0.908 & 0.054  \\  
 \midrule
        Method & Acc↓ & Comp↓ & C.R.↑ & Acc↓ & Comp↓ & C.R.↑ & Acc↓ & Comp↓ & C.R.↑ & Acc↓ & Comp↓ & C.R.↑ & Acc.↓ & Comp↓ & C.R.↑  \\ \midrule
        TSDF FT & 2.47 & 2.68 & 0.39 & 2.51 & 1.64 & 0.21 & 87.16 & 164.30 & 0.44 & 2.46 & 3.14 & 0.40 & 2.44 & 1.21 & 0.43  \\
        TSDF RS & 2.83 & 2.81 & 0.42 & 3.09 & 1.56 & 0.22 & 107.85 & 159.78 & 0.44 & 3.04 & 3.53 & 0.40 & 2.93 & 1.19 & 0.48  \\
        TSDF RRT\cite{schmid2020efficient} & 2.05 & 2.53 & 0.44 & 2.90 & 1.33 & 0.25 & 100.85 & 169.56 & 0.43 & 2.01 & 1.62 & 0.48 & 2.97 & 1.14 & 0.48  \\
        Offline FT & 1.09 & 1.02 & 0.70 & 1.77 & 1.21 & 0.66 & 1582.74 & 193.77 & 0.04 & 1.38 & 1.28 & 0.63 & \textbf{1.10} & 1.10 & 0.71  \\ 
        Online FT & 1.19 & 1.06 & 0.67 & 1.79 & 1.22 & 0.65 & 1506.59 & 202.52 & 0.05 & 1.37 & 1.30 & 0.62 & 1.19 & 1.09 & 0.71  \\ 
        Online RS & 1.88 & 1.20 & 0.57 & 1.53 & 1.10 & 0.70 & 652.66 & 211.00 & 0.27 & 2.62 & 2.25 & 0.37 & 1.89 & 1.11 & 0.66  \\
        \rs{Online even cover.} & \rs{1.21} & \rs{1.01} & \rs{0.74} & \rs{\textbf{1.15}} & \rs{1.08} & \rs{\textbf{0.77}} & \rs{54.24} & \rs{35.69} & \rs{0.48} & \rs{\textbf{1.50}} & \rs{1.29} & \rs{0.64} & \rs{\textbf{1.17}} & \rs{1.06} & \rs{0.76}  \\
        \rs{Ours Offline} & \rs{\textbf{0.96}} & \rs{\textbf{0.93}} & \rs{\textbf{0.76}} & \rs{1.31} & \rs{1.07} & \rs{0.74} & \rs{\textbf{31.66}} & \rs{\textbf{21.90}} & \rs{\textbf{0.73}} & \rs{1.51} & \rs{\textbf{1.19}} & \rs{\textbf{0.68}} & \rs{1.27} & \rs{\textbf{1.01}} & \rs{\textbf{0.76}}    \\       
        Ours & 1.07 & 0.95 & 0.74 & 1.29 & \textbf{1.04} & 0.75 & 48.48 & 34.16 & 0.60 & 1.37 & 1.20 & 0.65 & 1.43 & 1.02 & 0.74 \\
 \bottomrule
    \end{tabular}
    }
    }
    \vspace{-4mm}
\end{table*}
\vspace{-3mm}

NeurAR reconstructs scenes and plans the view path simultaneously, running parallel on two RTX3090 GPUs: the 3D reconstruction module, \ie optimization of NeRF on a GPU; the view planner module and the renderer module on the other one.  After a step of the view planner (also 700 iterations for the NeRF optimization), the optimization receives one or two images collected on the view path from cameras. We set the maximum views for planning to be 28 views, \ie 11k training iterations and about 15 steps for the optimization during the view planning. \rs{In each step, 1 or 2 views are selected along the planned path}. The network architecture of our NeurAR is shown in \figref{arc}. 

\noindent\textbf{Algorithmic Variants} To evaluate the efficacy of the proposed method, we designed baselines and variants of our method. For the implicit scene representation, we adopt TSDF as the baseline as it is oen of the most used representation for SLAM and autonomous reconstruction~\cite{schmid2020efficient,song2021view,isler2016information}.
For view path planning based on the proposed Neural Uncertainty, we construct two variants: one using a pre-defined circular trajectory with which existing work \cite{peralta2020next,song2021view} 
usually compares and the other one randomly sampling a viewpoint instead of using the view cost to choose NBV at each step. 

The variants are 1) \textbf{TSDF FT}: RGBD images and corresponding poses are collected from a \textbf{F}ixed circular \textbf{T}rajectory and are fed into the system 
sequentially. The voxel resolution of TSDF is 1cm. 2) \textbf{TSDF RS}: replace the fixed trajectory in \textbf{TSDF FT} with \textbf{R}andomly \textbf{S}ampled NBVs. 3)\textbf{TSDF RRT}: As for autonomous scene reconstruction most work is not open source, we re-implement the view cost defined in \cite{schmid2020efficient} which adopts TSDF representation. The online trajectory is planned with \textbf{RRT} and the view cost is defined according to the quality of a reconstructed voxel, measured by the number of rays traversing through it. Images and corresponding poses for the fusion are collected from the planned views and are fed into the system sequentially. 4) \textbf{Offline FT}:  The variant is NeurAR with views from the fixed trajectory and trained offline. 5) \textbf{Online FT}: The variant is NeurAR with views from the fixed trajectory.  6) \textbf{Online RS}: NeurAR with randomly sampled NBVs. 7) \textbf{Online even cover.}: \rs{NeurAR with views evenly distributed in a dome to cover the whole space.} 8) \textbf{Ours offline}: \rs{NeurAR trained offline, \ie trained from scratch with images precaptured from all the planned views from our proposed method. } 

\noindent\textbf{Metrics } The quality of the reconstructed models can be measured in two aspects: the quality of the rendered images (measure both the geometry and the texture) and the quality of the geometry of the constructed surface. For the former, we evenly distribute about 200 testing views 80m for $Alexander$, 3m and 3.4m from the center for other scenes, render images for the reconstructed models and evaluate PSNR, SSIM and LPIPS for these images. For the latter, we adopt metrics from iMAP~\cite{sucar2021imap}: $Accuracy$ (cm), $Completion$ (cm),  $Completion Ratio$ (the percentage of points in the reconstructed mesh with $Completion$ under a threshold, 30 cm for $Alexander$ and 1cm for others). For the geometry metrics, about 300k points are sampled from the surfaces.
\vspace{-2mm}
\subsection{Neural Uncertainty}


\figref{optimization} has quantitatively verified the correlation between Neural Uncertainty and the image quality. Here, we further demonstrate the correlation with examples. In this experiment, the reconstruction network is optimized by the loss defined in \eqref{logloss} with 73 images collected from cameras placed roughly at one hemisphere. \figref{lego} shows the images rendered from different viewpoints for the reconstructed 3D models and their corresponding uncertainty map during training. The uncertainty map is rendered from an uncertainty field where the value of each point in the field is $\frac{1}{log{\uncert}}$. Viewpoints of Row 1 are in the hemisphere seen in the training and the viewpoint of the last row is in the other hemisphere. At the beginning, for all viewpoints, the object region and the empty space both have high uncertainty. For the seen viewpoints, with training continuing, the uncertainty of the whole space decreases and the quality of the rendered images improve. When the network converges, the uncertainty only remains high in the local areas having complicated geometries. For the unseen viewpoint, though the uncertainty for the empty space decreases dramatically with training, the uncertainty is still very high on the whole surface of the object even when the network converges.

\figref{planning} shows the Neural Uncertainty can successfully guide the planner to planning view paths for cameras to look at regions that are not well reconstructed. For example, given current pose 15 and $\nerf$, a path is planned toward viewpoint 16 having a higher uncertainty map. The rendered image from viewpoint 16 from $\nerf$ exhibits poor quality (zoom in the image and the uncertainty map for details under the roof).

\subsection{View planning with Neural Uncertainty}

To show the efficacy of our proposed method, we compare metrics in \tableref{tablepsnr} and the rendered images of reconstructed 3D models using different methods in \figref{compare_view}. \tableref{tablepsnr} demonstrates except for the even coverage, our method outperforms all variants on all metrics significantly. Our proposed NeurAR achieves better reconstruction results with shorter view paths compared with other variants and existing work.

Compared with methods using the implicit representation without path planning (\textbf{Offline FT}, \textbf{Online FT}, \textbf{Online RS}), our method demonstrates significant improvements in the image quality and geometry accuracy. Methods without the planned views 1) are even unable to converge to a reasonable result in the large scene (NeRF will fail due to overfitting without carefully planned input viewpoint coverage), 2) have many holes in the objects as these regions are unseen in the images and 3) inferior image details on the surface of objects as these regions have a little overlap of different views, making inferring the 3D geometry hard (check red box in ~\figref{compare_view} for visual comparison). In addition, these variants tend to have ghost effects in the empty space. This is largely because the viewpoints are designed to make the camera look at the objects and the empty space is not considered. Notice that placing viewpoints covering the whole space without the aid of visualization tools as feedback is non-trivial for humans. 

\rs{Assuming we have the 3D shape of a scene with the target object in the center (which is our goal), to cover the whole space, we distribute viewpoints evenly at a dome centered at the scene center with a radius of 80 meters for $Alexander$ and 3 meters for others. All viewpoints look at the center. An even coverage path can provide a good reference as even coverage for many scenes of simple shape and texture is an optimum solution. Metrics for \textbf{Online even cover.}  and \textbf{Ours} in \tableref{tablepsnr} demonstrate that our planned views can achieve similar or even better reconstruction results and our method has no prior or only a minor knowledge about the scene.}

\rs{ \textbf{Ours offline} in most scenes gives better results than NeurAR trained online as using all views enforces multiview constraints at the start of the training. }

In addition to the lower mean PSNR shown in the \tableref{tablepsnr}, our method achieves much smaller variance regarding the PSNR of the rendered images from different viewpoints. For example, the PSNR variance of the rendered images for reconstructed cabins using \textbf{Online FT}, \textbf{Online RS} and our method is  32.55dB, 15.44dB to 4.78dB, indicating our method \rs{more even}.
Further, for the average path length in the smaller scenes, our NeurAR traverses 43.27m while \textbf{Online RS} about 70.24m; in the larger scene, our NeurAR traverses 907.60m while \textbf{Online RS} about 1329.75m.


For models using the reconstruction with TSDF (\textbf{TSDF FT}, \textbf{TSDF RS}, \textbf{TSDF RRT}), we render images via volume rendering. From the images in the last three columns in \figref{compare_view}, the surfaces exhibit many holes; in addition, the surfaces of the reconstructed models are rugged due to the noise in the depth images. Though NeurAR uses depth too, it depends on the depth images to accelerate convergence only at the early stage of training and decreases its effect after coarse structures have been learned. The finer geometry is acquired by multiview image supervision. NeurAR fills holes in \textbf{TSDF RRT} and provide finer details. It outperforms \textbf{TSDF RRT} in the image quality, geometry quality and path length (43.27m vs 57.39m). Though post-processing can be applied to extra finer meshes for the reconstructions using TSDF and get smoother images than our rendering images from TSDF directly, denoising and filling the holes are non-trivial tasks, particularly in scenes having complex structures.


    


        

\subsection{Ablation study}
\noindent\textbf{Training Iterations between Steps} The number of iterations for NeRF optimization $iter$ allowed for planning a view step affects final results.  We choose different iterations to run our NeurAR system in the scene $cabin$ and compare PSNR of the rendered images of the reconstructed models. PSNR for the models optimized using 300, 700, 1400 iterations is 26.16, 28.58, 26.91. 
\rs{This is because too few iterations may lead to uncertainty not being optimized well while too many steps may lead to overfitting to images added.}


\noindent\textbf{Depth noise} We construct variants of NeurAR using different noise magnitude from no noise, the noise magnitude of L515, to the double ($2\times \mu(z)$, $2\times \sigma(z)$ ) and the triple ($3\times \mu(z)$, $3\times \sigma(z)$ ) noise magnitude of L515. \tableref{tablenoerror} shows the metrics measuring the cabin models reconstructed by these variants: \rs{
NeurAR can be accelerated with very noisy depth at a cost of only a minor drop in the reconstruction quality. The robustness verifies the effect of dampening the depth supervision during the training in \secref{online} .}

\begin{table}[htpb]
    \caption{Influence of different noise magnitude }
    \tablelabel{tablenoerror}
    \centering
    \scriptsize
    \begin{tabular}{p{2cm}<{\centering}p{0.6cm}<{\centering}p{0.6cm}<{\centering}p{0.6cm}<{\centering}p{0.6cm}<{\centering}p{0.6cm}<{\centering}p{0.6cm}<{\centering}}
    \toprule
        Method & PSNR↑ & SSIM↑ & LPIPS↓ & Acc↓ & Comp↓ & C.R.↑\\
         \midrule
        Ours (no noise) & 28.71 & 0.926 & 0.043 & 0.92 & 0.90 & 0.75 \\
        Ours (with noise) & 28.35 & 0.902 & 0.062 & 1.07 & 0.95 & 0.74 \\
        Ours (noise x2) & 28.19 & 0.903 & 0.062 & 0.99 & 0.97 & 0.74 \\
        Ours (noise x3) & 27.65 & 0.900 & 0.063 & 1.00 & 0.94 & 0.77 \\
        \bottomrule
    \end{tabular}
    
\end{table}
\noindent\textbf{Number of views} \fl{To further study the influence, we set the maximum allowed views from 18 to 58 for the $cabin$ scene and provide the PSNR for the reconstructed models under different view settings.}
\vspace{-4mm}
\begin{table}[h!]
    \centering
    \caption{Evaluations on the reconstructed 3D models using different number of views}
    \scriptsize
    \begin{tabular}{p{2cm}<{\centering}p{0.6cm}<{\centering}p{0.6cm}<{\centering}p{0.6cm}<{\centering}p{0.6cm}<{\centering}p{0.6cm}<{\centering}p{0.6cm}<{\centering}}
    \toprule
        Views & PSNR↑ & SSIM↑ & LPIPS↓ & Acc↓ & Comp↓ & C.R.↑ \\
         \midrule
        18 & 22.47 & 0.802 & 0.155 & 2.18 & 1.31 & 0.59  \\
        28 & 28.35 & 0.902 & 0.062 & 1.07 & 0.95 & 0.74  \\
        38 & 30.03 & 0.919 & 0.053 & 0.90 & 0.86 & 0.77  \\
        58 & 29.39 & 0.926 & 0.050 & 0.82 & 0.90 & 0.79  \\
        \bottomrule
    \end{tabular}
    
    \tablelabel{numviews}
\end{table}

\section{Conclusion}

In this paper, we have presented an autonomous 3D reconstruction method based on an implicit neural representation, with the view path planned according to a proxy of PSNR. The proxy is acquired by learning uncertainty for the estimation of the color of a spatial point on a ray during the reconstruction. A strong correlation is discovered between the uncertainty and PSNR, which is verified by extensive experiments. Compared with variants with a pre-defined trajectory and variants using TSDF, our method demonstrates significant improvements. Our work shows a promising potential of implicit neural representations for high-quality autonomous 3D reconstruction. 

One limitation of NeurAR is the optimization of implicit neural representation is slow and the computation consumption is high. The optimization of the model between view steps takes about 50-120 seconds depending on the number of iterations, which is much slower than existing unknown area explorations. However, our method is agnostic to the underlying NeRF realization and therefore, we can easily swap the module for more advanced NeRF variants emerging lately~\cite{muller2022instant}.  Future work includes achieving acceleration of the training and rendering for NeurAR. 

The other limitation is that we assume known camera poses, which require extra location systems to make NeurAR work. Future work includes joint optimization of the camera poses, reconstruction, and view planning. Also, extending NeurAR to multiple agents for large scene reconstructions \fl{with distributed learning~\cite{huang2022tackling}} and \fl{adverse environments with the mmWave radar~\cite{chen2022mmbody}} are interesting directions.
\vspace{-1mm}

{\scriptsize
\bibliographystyle{IEEEtran}
\bibliography{IEEEabrv,egbib}
}

\newpage
\vfill
\end{sloppypar}
\end{document}